Antske Fokkens, Serge ter Braake, Niels Ockeloen, Piek Vossen,
Susan Legêne, Guus Schreiber, Victor de Boer

# BiographyNet:
# Extracting Relations Between People and Events

## Introduction

Humans are often called the smallest units of history. Any historical event can be traced back to the acts, decisions and thoughts of individuals. Simultaneously, these individuals share a lot of common character traits. They were born, have parents, most go to school, some have siblings, they died or will die and many of them have a career of some sort. The most famous individuals also have a 'claim to fame', which can range from being a famous politician to having murdered 27 people. The World Wide Web contains a massive amount of biographical data that begs to be analyzed with computational methods. Part of this biographical data is already (semi-)structured and is readily available for systematic computerised analyses. For this reason many projects analyzing biographical digital data have recently been undertaken.[1] BiographyNet is one of these projects.

BiographyNet is a digital humanities project (2012-2016) that brings together researchers from history, computational linguistics and computer science.[2] The project uses data from the Biography Portal of the Netherlands (BPN), which contains approximately 125,000 biographies from a variety of Dutch biographical dictionaries from the eighteenth century until now, describing around 76,000 individuals.[3] BiographyNet's aim is to strengthen the value of the portal and comparable biographical datasets for historical research, by improving the search options and the presentation of its outcome, with a historically justified NLP pipeline that works through a user evaluated demonstrator. The project's

---

main target group are professional historians. The project therefore worked with two key concepts: "provenance"—understood as a term allowing for both historical source criticism and for references to data-management and programming interventions in digitized sources; and "perspective"—interpreted as inherent uncertainty concerning the interpretation of historical results.

This paper provides an overview of this project and describes the state-of-affairs of the project as of March 2016. Like most other 'digital humanities' projects, BiographyNet was set up as an interdisciplinary endeavour. BiographyNet was envisioned by researchers from history, computational linguistics and computer science. Bringing these disciplines together should result in a demonstrator that can mine interesting facts from text, help the historian reason over the data and provide a user-friendly interface to help answering historical questions. The main idea of the project was to go beyond the basic search principles of the Biography Portal of the Netherlands, and to generate relationships between peoples and events, geographical movements of and networks between people from the Biography Portal and to see what they tell historians about the formation of Dutch society and the 'boundaries of the Netherlands'.

The rest of this paper is structured as follows. We will start with an outline of the general background and motivation for the project in Section 2. Section 3 provides a more detailed overview of the data provided by the BPN and the first outcome of data analyses. This section also addresses the limitations inherent to the data. This is followed by a description of the kind of historical questions that can be addressed using this line of research in Section 4.

The rest of this paper describes the methods and models that are (being) developed as part of the project. We first describe the natural language processing tools that were used to identify information automatically in Section 5. Our data model is described in Section 6. This covers both how we model the overall outcome of our analyses in RDF and how we address main methodological issues through provenance modeling. We then describe the visualizations we designed for our system. Finally, we conclude and describe future work within and beyond the proposal.

## 2. Background and Motivation

Since the nineteenth century especially, biographical dictionaries have been compiled in most countries, often hand in hand with a new sense of national consciousness. Biographical dictionaries contain short, often semi-structured biographies on people deemed noteworthy. They contain the essence of 'normal' hundreds of pages biographies, but avoid too much context or broader historical reflections. Information about other people than the biographee is also omitted for the most part. Biographical dictionaries therefore are mostly useful as reference works and do not provide much insight into any historical period. The sheer quantity of biographical data, however, gives ample opportunity for quantifications.[4]

The digital age has changed the way academics work in every discipline. Computers can process digital data much faster than humans can do, they are able to show patterns and statistical analyses and can detect links that otherwise would be hard to collect. Data from biographical reference works offer themselves for computational analysis since they tend to use a common format, having emerged as a historiographic genre in the 19th century that has remained largely constant up until the present day. Individuals that have been included in the reference works share a set of common characteristics, such as having a birthdate, a partner, a profession, and a network. They share these similarities with other historical figures that appear in other online-resources. Within specific biographical sources they furthermore share characteristics (for instance: same profession, same gender, same claim to fame). These biographical characteristics can be relatively easily identified by a machine. With tools and approaches from the digital humanities, BiographyNet aims at both quantitative analyses of such (linked) data, while also providing leads for more qualitative research questions. The project has explored the extent to which linked data with an NLP enrichment and state of the art visualization can bring biographical data to a higher plan for historical research.

## 3. The Sources: the Biography Portal of the Netherlands

BiographyNet is built upon the Biography Portal of the Netherlands (BPN). All enrichments and visualizations take digitized data from the BPN as its input. It follows that the potential research that can be carried out within the project directly depends on the data that is provided by the BPN: the biographical dictionaries, their selection criteria, the research carried out in writing biographies and their format directly determine which questions can be addressed with this

---

data and which cannot because the input data does not contain the necessary information or is biased. In this section, we describe the dataset. We analyze the main properties of the data and critically reflect on the selection policies of various sources pointing out both the potential and the limitations of the data.

The Biography Portal of the Netherlands (BPN) contains the biographies of 23 different biographical resources (most of them biographical dictionaries) from the eighteenth to the twenty first century. A short description of each resource can be found on the website of the BPN.[5] The dictionaries are collective biographies in the sense that they contain the lives of a group of people who share certain characteristics: being a famous socialist, being a famous woman, or simply being famous. The fame factor was always the determining factor as to whether a potential biographee was included, with Elias' Vroedschap van Amsterdam being a notable exception (all members of the City's Council were included).[6] Of course 'fame' is a rather vague, subjective and unquantifiable criterion and therefore we are dealing with sources that are highly biased in their selection.[7] In order to highlight these biases we made some analyses of the sources ourselves, which you can find in Figure 1. The numbers are based on the (incomplete) categorizations that the people who created the BPN added as metadata.

If we take the three grand dictionaries that are supposed to cover all categories of people, Biographisch Woordenboek der Nederlanden (VDAA),[8] Nieuw Nederlandsch Biografisch Woordenboek (NNBW)[9] and Biografisch Woordenboek van Nederland,[10] comprising a total of 46,760 biographies (respectively 23,587, 21,096 and 2,077), then we see some clear biases.[11] For example, the NNBW is considered to be more reliable than the VDAA, but it is far more biased containing an enormous number of religious people, mainly protestant clergymen born in the seventeenth and eighteenth centuries. The BWN, only dealing with the years after the NNBW, clearly shows a shift in categories, with a strong preference for politics, literature and science. It is more balanced than the NNBW, but not more than VDAA. In general, more people from modern centuries are described than those living in previous centuries.

---

5   http://www.biografischportaal.nl/about/collecties

6   http://resources.huygens.knaw.nl/retroboeken/elias/#page=0&size=800&accessor=accessor_index&source=1

7   Serge ter Braake / Antske Fokkens, How to Make it in History. Working Towards a Methodology of Canon Research with Digital Methods, in: S. ter Braake, A.S. Fokkens, R. Sluijter, T. Declerck and E. Wandl-Vogt (eds)., *Biographical Data in a Digital World. Proceedings of the First Conference on Biographical Data in a Digital World 2015*. Amsterdam: CEUR online proceedings 2015, 85–93.

8   http://resources.huygens.knaw.nl/retroboeken/vdaa/#view=imagePane, latest retrieved 15 March 2016.

9   http://resources.huygens.knaw.nl/retroboeken/nnbw/#view=imagePane, latest retrieved 15 March 2016.

10  http://resources.huygens.knaw.nl/bwn, latest retrieved 15 March 2016.

11  Please note that the metadata from VDAA are quite incomplete compared to those of most other sources, which could have a relevant impact on the results provided here.

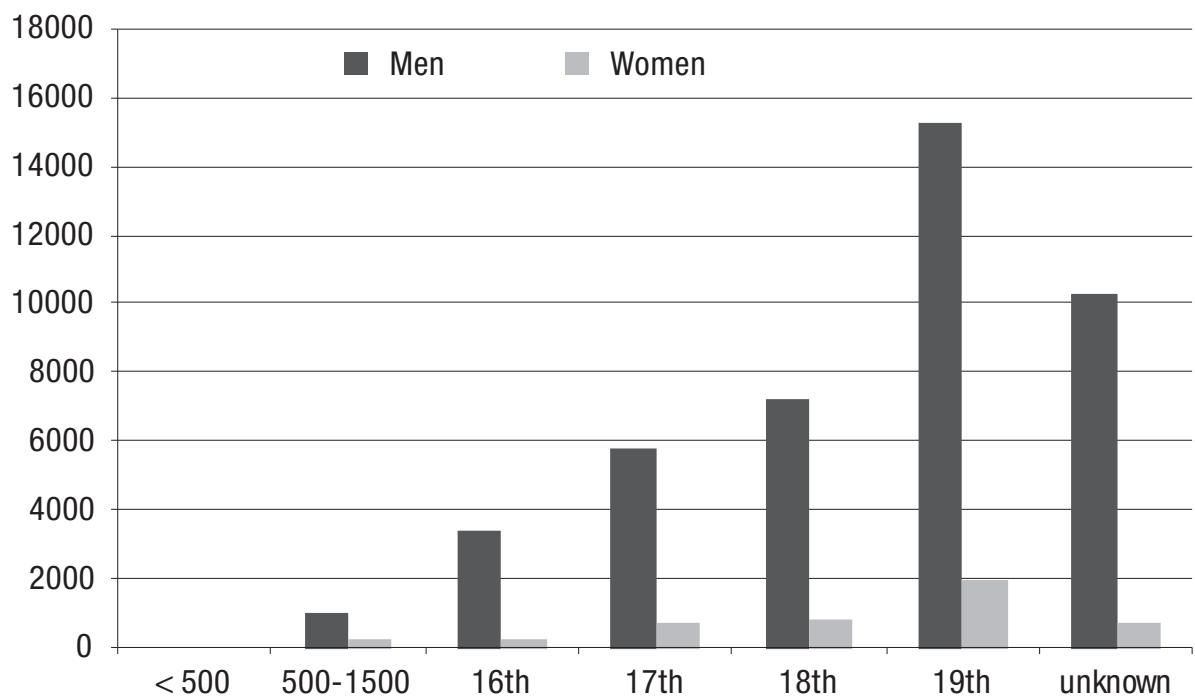

Fig. 1: Total men and women in BPN over the centuries[12]

## Problems of representation

For any editor of a biographical dictionary it is difficult to set effective entry criteria. In any reference work there seems to be an overrepresentation of certain categories of people and an underrepresentation of others (most notably women). Ever since the first biographical dictionaries were published, their editors have been criticized for (not) including certain people. Despite great efforts from their editors to justify their choices, it has become abundantly clear that certain biases cannot be avoided completely.[13] Guidelines on noteworthiness are

12   Note: people from the twentieth century were left out, because some sources (mainly VDAA) date from before this period.

13   E.g. Hendrik F. Wijnman, Project voor een vervolg op het nieuw Nederlandsch Biografisch Woordenboek. In: *het jaarboek van de maatschappij der Nederlandse Letterkunde te Leiden (1962-1963),* Leiden: Brill, 1963; Léon Hanssen, 'Op zoek naar een onbekende. Biografische lexicons als wetenschappelijk hulpmiddel', *Biografisch Bulletin* 5, 1995, nr. 1, 77–83, at 78; Ira Bruce Nadel, *Biography. Fiction, fact & form*, London and Basingstoke: Palgrave Macmillan, 1984, at 52.

helpful to select the people for inclusion, but always leave plenty of room for subjectivity. In the digital age, Wikipedia is famous for documenting their criteria well and for keeping a lively discussion going on this subject.[14] The criteria for who is deemed noteworthy has also change over time. Entire categories like sportspeople—who were largely irrelevant at the turn of the century—and television personalities are now included. The editors of the BWN provided interesting self-reflections on their selection policy. In Volume V no less than 42% of the people included came from Amsterdam, The Hague, Rotterdam and Utrecht. According to the editors this imbalance was unavoidable since those cities formed the political, economical and cultural heart of the Netherlands.[15]

A problem of a more general nature is the lack of women. All dictionaries that are not particularly focused on women, such as the Digitaal Vrouwen Lexicon, contain few descriptions of women's lives.[16] The BWN only dedicated 7% of their biographies to women, the VDAA only 2.7% (of which 30% were poets), and the NNBW only 2.1% (of which 29% were religious women).[17] In general there are a significant number of medieval women depicted, most likely because of the large number of saints and female royalty. Even though the Vrouwen Lexicon at least partially addresses this gender bias, it confirmed the biases of its predecessors by putting all women from the VDAA and the NNBW on their 2004 longlist of women who merited description.

Another bias is the length of the entries. If someone merits inclusion in a biographical dictionary, how many pages does he or she deserve? The editors of VDAA defended themselves against allegations of a distorted relation between the length of certain entries and the importance of the biographee by stating that more text was dedicated to some people because there was little known about them previously. The editors of the NNBW stated that a certain discrepancy was unavoidable, since they did not want to restrict their authors too much with rigid guidelines. The editors of the BWN had to admit that in the course of the project they changed from very factual biographies to biographies that permitted striking details or interesting facts that illustrated the character of the

---

biographee. This means that the people who were included at a later stage of the project have relatively extensive biographies for reasons unrelated to their note-worthiness.[18]

The question is wether these biases and unfairness in the biographical dictionaries pose a problem for the researcher. We dare to say it does not, as long as the researcher remains conscious of the limitations. Historians always have to make do with an incomplete and non-representative collection of resources. The reconstruction of the past is always a distorted image of what really happened, because not everything was documented or even could be documented and because historians always bring along their own (subjective) perspectives when composing a narrative. The biases in the BPN therefore pose no new challenges to historians and if we approach these biographies as big data, then computational techniques can even help us to understand these biases. Of course one could say that by using this collection of dictionaries only old and badly substantiated canons of Dutch history are repeated. This would be the case if researchers were to exclusively consult the BPN, which no serious researcher would ever do. As long as researchers remain aware of the richness of the sources outside the BPN, they can make good use of it and the tools developed as part of BiographyNet. In the next section, we outline which kind of research questions can be addressed with this data (and the BiographyNet tools) when taking the data's bias into account.

## 4. Historical Questions

The data of the BPN is rich and can be used to answer all kinds of questions. We will divide them here in prosopographical questions, thematic questions and historiographic questions.

### Prosopography

Simply stated, a prosopography is an analysis of the measurable characteristics of a well-defined group.[19] We could, for example, take the lives of all councillors of the Council of Holland in the first half of the sixteenth century and analyze their age when first appointed, their education, the social groups they belonged to, where they were born et cetera. By mapping such group portraits, we can see who was appointed, how this changed over time and what influence that may have had on the way the institution worked in its intermediary role between the prince and the subjects.[20]

---

18   *Biografisch Woordenboek van Nederland V*, vii-viii. See also *Biografisch Woordenboek van Nederland N* VI, ix; Jos Gabriëls, 'Portretten in miniatuur: het Biografisch Woordenboek van Nederland' in: B. Toussaint and P. van der Velde (eds), *Aspecten van de historische biografie*. Kampen: Kok Agora, 1992 50–64, at 55–60.

19   A nice summary of all aspects is provided in: Katharine Keats-Rohan, *Prosopography approaches and applications: A Handbook*, Oxford: Occasional Publications UPR, 2007.

We could 'prosopographize' all sorts of groups, like the painters in the first half of the seventeenth century, students at the University of Oxford in the Middle Ages or the nobility in sixteenth century Flanders. Since biographical dictionaries also contain information on so-called second rate famous people in history, they often provide more useful data than, for example, DBpedia and other general encyclopedic works. Naturally, this still does not mean that all people who ever made some kind of mark on history are present in the Biography Portal of the Netherlands. The kind of group under investigation largely determines wether the BPN is a suitable (and sufficient) resource for the investigation. When dealing with people who have a profession or characteristic that guarantees inclusion in one of the dictionaries, it forms a rich resource that provides, at the very least, a solid basis. When the target group shares a general feature that will not automatically lead to enough fame, the resource will often prove too limited. For instance, historians analyzing all students from Leiden University will not find the majority in the Portal, whereas people analyzing all governors-general from the Dutch Indies *will* find everyone there.

Other than being a useful research method in its own right, prosopography is important as it provides the context for an individual in a biography. It is, for example, difficult to say anything about a person (well educated, ingenious, ill tempered) without comparing them to others operating in the same context. Very often however, individual biographies lack such 'supporting' prosopographical analyses, because their compilation is very time consuming.[21]

One of the challenges BiographyNet has faced is enabling quick comparisons between people and (the early stages of) prosopographical analyses by extracting basic information on all entries in the portal. The metadata of the Portal already provides some basic information such as date and place of birth, but structured information on marriage, parents, religion and appointments in offices is often lacking. We manually carried out a prosopographical survey of the

---

71 governors-general of the Dutch Indies between 1610 and 1949 to establish what is necessary in order to manually identify this information for a specific use case. It took us two weeks to extract all the relevant information from the biographies and analyze the data in graphs. In Section 5, we will describe the NLP (Natural Language Processing) tools we developed to aid this kind of research. The often unstructured format of the biographies, and their heterogeneous provenance, make it a non-trivial task to do this well. Even though we are aware that the NLP does not produce (near) perfect results, historians should be able to see general patterns to help them formulate hypotheses and further research questions.

### Thematic Questions

The previous example illustrates research that can be performed more quickly with the help of computational techniques. We also aimed to address more intellectually challenging questions that are practically impossible to answer without computational methods, because the historian would not know where to start looking in the data. One question which may be addressed is for example the involvement of Dutch elites with the 'discovery' and exploration of the *New World*. After 1492, when Columbus first reached the Americas not much changed in the Low Countries at first, except for some minor remarks in the chronicles. Gradually however, Dutch elites became involved with the process of colonial trade, colonization and exploration. To detect this process, one would need to dig deep into the available texts to search for instances of American place names, American products, international ships, overseas jobs et cetera. Without a computer a person would have to read all available biographies on people living after 1492 to trace this "Americanization" of Dutch elites, which is a very undesirable endeavour. Our demonstrator can search for key phrases, detect new word occurrences and trace *America* in all data.

In order to do this successfully the system needs to be queried for concepts and locations related to America in the sources. Even though the original search interface of the BPN supports key word search, the possibilities of investigating the occurrence of concepts is limited. BiographyNet extends this in three ways. First, our NLP analyses tag the lemmas occurring in a specific biography, so that the search of a specific noun will identify both singular and plural occurrences and verbs can be identified independently of the form (person agreement or tense) they occur in. Second, we tag content words with WordNet identifiers. This way we can search for synonyms of terms. Finally, the current BNP interface simply returns a list of terms that mention a specific word. In the structured data created in BiographyNet, users can extract statistics on concepts including correlations between a concept and specific sources, the concept and the date of birth (or death) of the people in whose biographies they occurred, etc. Geographical locations are particularly tricky, because colonies received the names of already existing places in Europe (New Amsterdam, New York, New Orleans). Heuristics that look at other locations mentioned in the text (other place names, titles of officials, et cetera) can ensure that our tool is able to make correct identifications.[22] One of the major challenges of this project was to create a tool that is both sophisticated and user friendly enough to be used by a historian to answer this and similar questions.

**Historiographic Questions**

Biographical dictionaries do not only give us building blocks to make reconstructions of the past, they also contain constructions that were built in the past and provide insights into the way history was written in a particular period. Biographical dictionaries and other encyclopedias are well suited for historiographical research, since they, in the words of Manfred Beller, 'summarize received knowledge and thus transmit discourse from specialized cultural fields into the domain of public opinion.'[23] The BPN provides very interesting material for such research, because it contains biographical data compiled from the eighteenth century until the present day.

A relatively straightforward task would be, for example, to see how many adjectives are used in every dictionary compared to the entire number of words used. Dictionaries that use more adjectives are almost by definition more subjective than others. The academy-student project *Time Will Tell a Different Story* aimed specifically to automatically find differences in the way people are described.[24] Many 'heroes' of the past have been judged differently over time, especially when they are war heroes or politicians.[25] It is also interesting to see what kind of adjectives are used in biographies from different time periods and biographical dictionaries and to see the developments there.

---

22  Serge ter Braake / Antske Fokkens / Fred van Lieburg, Mining Ministers (1572-1815). Using Semi-structured Data for Historical Research, in: L.M. Aiello and D. McFarland (eds), *6th International Conference on Social Informatics (workshops)* Barcelona: Springer, 2014, 279–283.

23  Manfred Beller, Encyclopedia, in: M. Beller and J. Leerssen (eds), Imagology. *The cultural Construction and literary representation of national characters. A critical survey,* Amsterdam and New York: Rodopi, 2007 319–323, at 322.

24  http://www.biographynet.nl/uncategorized/time-will-tell-a-different-story-presents-at-dhb-2015-and-network-institute-annual-event/. Miel Groten / Serge ter Braake, / Yassine Karimi / Antske Fokkens, Johan de Witt. De omstreden republikein, *Leidschrift* 31, 2016, vol 2, 59–73.

### Future Questions

BiographyNet works with Dutch data on (mainly) Dutch people and in the Dutch language. However, borders of European states and nations were fluctuated greatly from the Middle Ages until at least the nineteenth century, and within the Netherlands, Dutch was not the only language used. It would be of great interest to be able to apply the BiographyNet software to biographical data in other languages as well. There are two topics in particular which would attract our attention: 1) To attempt to evaluate how people of international renown, such as Erasmus, Voltaire, Descartes and Vincent van Gogh, are described in dictionaries from different countries. Are there big differences and how can we explain them? 2) To attempt to make general historiographical comparisons addressing questions like: How are people generally described in different countries? With what judgments and properties? Were Dutch historians more nationalistic than their German counterparts? Did the French ascribe different merits to their heroes than the Italians? Such questions have intrigued historians for a long time, but until now, no computational research has been carried out that may lead to possible answers. Even though several NLP tools used in BiographyNet can only be used to analyze Dutch, the methodology is language independent and the principles can be applied to biographical data written in other languages. We hope to continue this line of research in future projects and collaboration with researchers from other countries.

## 5. Data enrichment through NLP

As pointed out above, the biographical resources of the BPN are accompanied with metadata, but this metadata is often very limited. Most information is presented as text. One of the main goals of BiographyNet is to analyze these texts using Natural Language Processing (NLP) tools and represent the infor-mation contained in the text as structured data in RDF.[26] In this section, we describe the process of going from text to structured data using NLP.

---

25   For example governor of the Dutch Indies Jan Pietersz Coen, whose statue in Hoorn still is subject of public debate: https://nl.wikipedia.org/wiki/Jan_Pieterszoon_Coen

### Goals

In previous work,[27] we have used the metaphor of a house to illustrate how historical narratives are structured. The idea is that historians build their analysis using data at their disposal; they construct their narratives by putting the facts they found in their sources together with established historical methodologies. If we see a narrative as a house, the data and facts are the building blocks. The goal of the NLP analyses is to identify such building blocks in text, which can be used by historians to create new buildings.

Based on the three types of historical questions outlined in Section 4, we can distinguish three kinds of information. Prosopographical studies mainly need access to events in people's lives: place and date of birth, what they studied (and where), when they were appointed to a specific position and what they did before that. For thematic questions, we want to extend the possibility of identifying concepts beyond key-word search. This means that we want to be able to identify words regardless of their surface form (lemmatization) and of the exact word (synonym) chosen to express it. Historiographic research benefits from information on the sentiment that occurs in biographies: are the attributes associated with a person positive or negative?

Even though each line of research has its own focus, information relevant for one research domain is often also relevant for other domains. For instance, the events considered relevant by individual sources can be of interest in a historiographic study and the judgment or sentiment associated with a specific group of people can be addressed in prosopographic work.

Taking these domains of interest into consideration, we aim to detect the following building blocks with our NLP pipeline:

- Events with their participants, time and location

---

26  Jeremy J. Carroll / Graham Klyne, Resource description framework (RDF): Concepts and abstract syntax. W3C recommendation, W3C, 2004,  http://www.w3.org/TR/2004/REC-rdf-concepts-20040210/
27  Antske Fokkens / Serge ter Braake / Niels Ockeloen / Piek Vossen, /Susan Legêne / Guus Schreiber. BiographyNet: Methodological issues when NLP supports historical research, In:*proceedings of 9th International conference on Language Resources and Evaluation (LREC 2014)*. Reykjavik: European Language Resources Association (ELRA), 2014

- Term analysis: lemmatization and word sense disambiguation (identifying synonyms)

- Opinion mining and sentiment tagging

We developed a pipeline of NLP modules that aim to cover all these tasks. The next subsection describes the NLP pipeline. This is followed by a description of our approach to translate the outcome of our NLP analyses to structured data.

### NLP pipeline

Figure 2 provides an overview of the NLP pipeline used for Dutch. The blue boxes represent the various modules of the pipeline. The orange boxes indicate the in- and output data of each module. The in- and output data is represented in the NLP Annotation Format (NAF).[28] NAF is an extensible layered XML format specifically designed to support interoperability between NLP tools. We will provide a brief description of the modules and their purpose in this section.

The first step in our processing pipeline is tokenization. This is carried out by the ixa-pipe-tok module,[29] a rule based module which uses punctuation and lists of abbreviation to identify tokens and sentence boundaries in various languages. The tokenized text is then fed to the Alpino parser.[30] The Alpino parser provides terms: each term represents a word consisting of one or more tokens (e.g. verbs and their particles are considered to be one term, even if they are separate tokens). Alpino also provides the lemma, Part-of-Speech (PoS) and more fine-grained morphological information of the term. It furthermore outputs the constituent structure of the sentence as well as the syntactic dependencies. As mentioned above, lemmas can be used to identify words occurring in different surface forms (abstracting from plural marking on nouns, adjective endings or verb conjugation), but the rest of the information provided by these first two modules is not directly relevant for the historian: they provide the necessary input for modules aiming at semantic interpretation further down the pipeline. Heideltime is a pattern and rule based module that identifies and nor-

---

malizes time expressions.[31] The tokens and terms are also input of the term tagger. This tagger takes a lexical resource that relates lexical entries to identifiers in an ontology as input and aims to identify occurrences of these lexical items by looking at tokens and terms. In BiographyNet, the term tagger identifies mentions of professions as well as family relations.

The ixa-pipe-nerc module aims to identify named entities and classifies whether they refer to a person, organization, location or other entity (miscellaneous). Ixa-pipe-ned module aims to disambiguate the names that are identified by ixa-pipe-nerc. It uses DBpedia spotlight to link named entities to entities in DBpedia representing a specific person, organization or location.[32] The DBpedia-spotlight-cltl also aims to identify entities in text and link them to DBpedia, but it aims to perform both steps in one go. As such, it obtains higher recall (since it will also consider strings that the named entity recognizer missed), but lower precision. VUA-DSC-WSD performs word sense disambiguation. It tries to identify the most likely meaning for a given word given the context. Word senses are represented using WordNet identifiers. WordNet is a taxonomy that represents word meanings by defining sets of synonyms (so-called synsets) which are organized in a hypernym-hyponym hierarchy.[33] Words with the same meaning will receive the same synset identifier. This step thus does not only provide disambiguation of specific terms, but also the means to identify synonyms. WordNet identifiers can be linked to representations in other resources using the Predicate Matrix.[34] The vua-pm-tagger maps Word-Net identifiers to FrameNet frames.[35] FrameNet is an English lexical database that represents events and their participants. For instance, FrameNet Frame for the Education_Teaching includes the following roles: subject (the subject of study), establishment (the school, college or university), student (the instructed

---

person), teacher (the instructor). Each frame is linked to lexical items that evoke this frame, e.g. words such as learn, teach, study can all evoke the Education_Teaching frame. The Dutch Predicate Matrix maps Dutch WordNet identifiers to English lexical items which are in turn related to FrameNet frames.

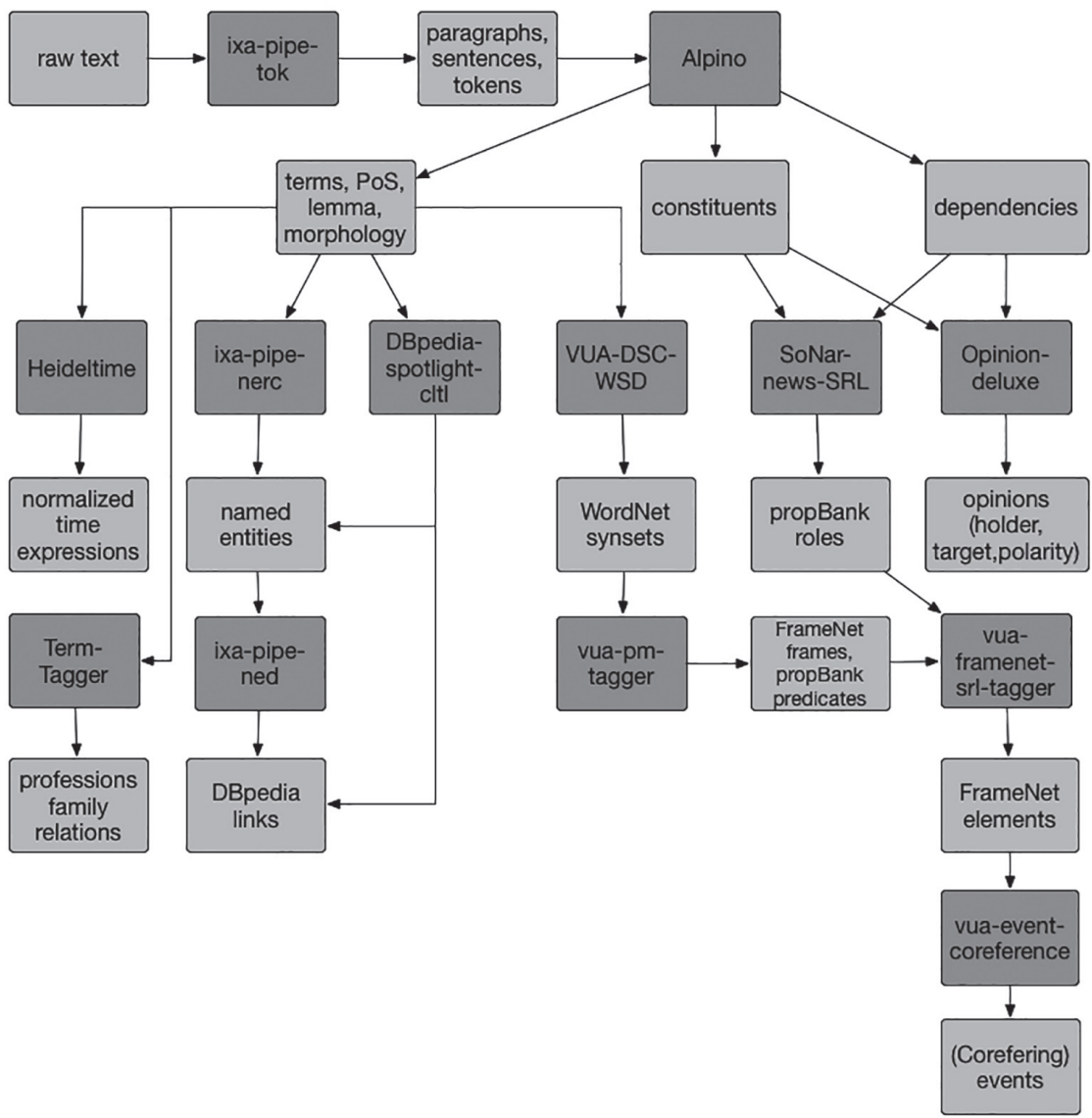

Fig. 2: NLP pipeline used in BiographyNet

The SoNar-news-SRL module identifies which syntactic argument plays which semantic role.[36] It is trained on SoNar, where abstract semantic roles in the form of Arg0, Arg1 based on propBank have been annotated on top of Alpino's syntactic analysis. The Predicate Matrix can map these abstract propBank roles to Frame Elements. The vua-framenet-srl-tagger indicates which argument of the verb study is the student, which is the subject of study and which is the institution. Finally, the event coreference module tries to identify which expressions refer to the same event. It achieves this by comparing the type of event (the meaning in WordNet and FrameNet) and other known information, such as the participants involved, the time and place of the event. However, biographies rarely mention the same event twice and this module mainly outputs singletons. Finally, the opinion minor identifies which expressions express a positive or negative opinion as well as the holder of this opinion.

### From Linguistic Annotations to Structured Data

The output of the NLP pipeline is a NAF structure that contains the events mentioned in the biography together with the event's participants and their semantic roles (i.e. what role they had in the event). The semantic role labeller also relates the locations and temporal indications that are mentioned to the event. In addition to events, the NAF structure indicates which concept (represented as a WordNet identifier) a term refers to and contains a list of professions and family relations mentioned in the text. Once a biography has been passed through the NLP pipeline and all this information has been identified, we apply an additional step to interpret these linguistic annotations and translate them to RDF structures. This procedure of creating a new structured biography based on the NLP analysis consists of four steps.

The first step carries out direct translations of information in NAF to RDF. In this step, all concepts (WordNet identifiers), lemmas and FrameNet frames identified by the NLP pipeline are linked to the biography. In the second step, we extract the events and participants from NAF. In this step, we combine in-formation from several modules to come to the final interpretation. We first extract predicates and semantic roles as provided by the SoNar-news-SRL and vua-framenet-srl-tagger. We then check each of these roles to see whether they are identified as a person, location or organization by the named entity recognizer or whether Heideltime recognized them as a temporal expression. In these cases, we verify whether the span of the semantic role and the interpretation is correct. If there is a mismatch, we prefer the output of the named entity recognizer or Heideltime over the SRL output, i.e. if the SRL output indicates a given argument is a locative, but it contains a temporal expression according to Heideltime, we will assign a temporal relation between the event and argument in our final output. The third step involves interpreting references to professions and family relations. For each term that indicates a profession or family relation, we interpret the syntactic patterns to establish who has the profession or who the family member is. For instance, in the sentence *her father was a doctor,* we link *doctor* to *father* and *her* to *father.* In the sentence, *her son Frederik was born,* we link *Frederik* and *her* to *son.* In the fourth step, we interpret pronominal references. If the biography's subject is female, the *her* in the aforementioned examples will be linked to this subject in this step.

---

36  The SoNaR-news-srl is a reimplementation of the system described in Orphée De Clercq / Veronique Hoste / Paola Monachesi. Evaluating automatic cross-domain Dutch semantic role annotation, In: *proceedings of 8th International conference on Language Resources and Evaluation (LREC 2012)*. Istanbul: European Language Resources Association (ELRA), 2012. The reimplementation was created as part of NewsReader. See: Piek Vossen / Rodrigo Agerri / Itziar Aldabe / Agata Cybulska / Marieke van Erp / Antske Fokkens / Egoitz Laparra / Anne-Lyse Minard, Alessio Palermo Aprosio / German Rigau / Marco Rospocher, NewsReader: Using knowledge resources in a cross-lingual reading machine to generate more knowledge from massive streams of news, *Knowledge-Based Systems* 110, Elsevier: 2016, 60-85.

All information extracted in these four steps is linked back to the original tokens in the text using the Grounded Annotation Framework (GAF).[37] We thus represent the exact location of where specific information was mentioned. This allows end-users to go back to the original source and verify wether the automatic interpretations are correct. We can also add information about the certainty of statements and the sentiment associated with the chosen words to the tokens. We plan to add a fifth step to our interpretation algorithm in the near future that includes this information in our model. This step finalizes the NLP analysis and interpretation algorithm within the BiographyNet project. In Section 6, we will describe the data model in more detail.

### Evaluation

The evaluation of the NLP analyses is currently ongoing. In this subsection, we will address the setup of our evaluation and what needs to be taken into consideration while evaluating NLP analyses that provide input for further re-

---

search. One of the most important aspects of the evaluation is that it should raise awareness to the end user about what the NLP analysis can do and what they cannot do. As we will argue, the standard precision and recall evaluations are not sufficient to provide the necessary insights to historians using the output of our automatic analyses. It is also important to provide insight into the kind of errors made by analyses, so that end users are aware of potential biases introduced by the tools. We will first explain this idea as outlined in previous publications and then outline how we are trying to tackle this issue in BiographyNet.

It is a well-known fact that automatic text analyses do not yield perfect results. One of the main challenges in using the outcome of such analyses in digital humanities projects lies in determining whether the results are reliable enough to be used to draw conclusions. This question is particularly challenging, because it cannot be answered by simply looking at the overall performance of the tools that are used, it actually depends on the kind of errors that the tool makes and on the research question. We will illustrate this using an example from previous publications and a study carried out as part of our collaborative research in BiographyNet.[38]

We mentioned above that disambiguating locations is one of the challenges in BiographyNet. Since the BPN only includes Dutch people or people who were residents of the Netherlands for a significant part of their lives, an algorithm always selecting the location that lies in or is closest to the Netherlands will yield very high accuracy. Whether the couple of percent of locations are problematic or not, however, depends on the historian's research question. If an historian investigates where officials in The Hague were born for instance, these errors will have little to no impact in identifying the region most officials came from. If, on the other hand, the question at hand concerns officials working in the overseas territories, the algorithm is highly problematic: many places in former colonies were named after cities in the home country and all of them will be misclassified by the algorithm.

---

38   See: Antske Fokkens / Serge ter Braake / Niels Ockeloen / Piek Vossen / Susan Legêne / Guus Schreiber. BiographyNet: Methodological issues when NLP supports historical research, in: *proceedings of 9th International conference on Language Resources and Evaluation (LREC 2014)*. Reykjavik: European Language Resources Association (ELRA), 2014,. and Niels Ockeloen / Antske Fokkens, / Serge ter Braake / Piek Vossen / Victor de Boer / Guus Schreiber / Susan Legêne, BiographyNet: Managing provenance at multiple levels and from different perspective, In: *Proceedings of the 3rd International Workshop on Linked Science-Volume 1116*, 59–71. Sydney: CEUR online proceedings, 2013.

Another issue to take into consideration is that while historians typically show a preference for methods that achieve high precision, high recall is still more important for certain research questions. The canonization study we carried out in BiographyNet forms a clear example of such a situation.[39] In this study, we aim to identify who is mentioned most often in the data. If we use the outcome of the named entity recognizer, most mentions we extract will indeed be person names. However, it is very well possible that the algorithm will fail to recognize certain person names resulting in the risk that we completely miss out on certain persons occurring in the corpus. Rather than using the named entity recognizers output we therefore made use of a basic pattern matching algorithm, which retrieved and counted any string that remotely looked like a name. The algorithm also counted many strings that are not names at all, but the historian studying the outcome could see at a glance that these strings were not names and discard them in the final analysis. The presence of errors (low precision) thus does not have a negative impact on the outcome of the historian's research question, whereas an approach with higher precision at the cost of recall may have.

The examples above illustrate that the kind of errors a module makes and the research question are both of importance while determining if the results of automatic text analysis can be used to provide reliable answers. It is not feasible to carry out analyses that would identify potential biases for any possible study historians might want to carry out with the data. We therefore aim to carry out evaluations that raise awareness about possible errors and biases to end users and that illustrate how historians can design evaluations that verify whether errors in the analyses directly influence their research results. We achieve this by carrying out two kind of evaluations: intrinsic evaluations of the tools and extrinsic evaluations of the possibility of using these tools to answer specific research questions. For the intrinsic evaluation, we manually annotate biographies and calculate the tools precision and recall on identifying named entities, time expressions, events and semantic roles. For extrinsic evaluation, 25 historical questions were designed that cover various studies and question types. Some historical questions have been answered through manual analysis, such as the aforementioned prosopographical study on governors of the Dutch In-dies. In this case, we can verify whether the output of our NLP pipeline leads to the same trends and observations as the manual analyses. It is, however, not always possible to verify all relevant data by hand. For cases where the amount of data is too large, we apply a method that we have dubbed hypothesis-based sampling. This method aims at verifying how reliable the analyses are for verifying the historian's hypothesis on the research question. For instance, if the historian hypothesizes that the 19th century resource Van der Aa is more subjective than the early twentieth century resource the NNBW, the exact accuracy on identifying subjectivity is not the most relevant for a reliable outcome. Instead it is more important that errors in classification are equally distributed over both resources. For such a study, we evaluate the reliability of the tool based on three samples: one selection of data that confirms the hypothesis, one that would reach the opposite conclusion and one sample that would lead to similar subjectivity between the two resources. The evaluation sample would thus include texts that were judged highly subjective, average subjective or not subjective at all from both resources. If manual analysis leads to the same conclusions on these samples as the tools, this provides a strong indication that the overall outcome is reliable. By verifying data from all three categories in both resources, we reduce the risk that the automatic analysis systematically misses or overclassifies subjective terms in one of the resources and influences the overall outcome of the hypothesis without the end user knowing this.

## 6. Data model

The data model we use can be divided in two main components: (1) the overall architecture that links information from biographies to the person and models the provenance of information and (2) the data model used to model biographical information itself. We will describe both parts in this section.

### The BiographyNet Schema

Ockeloen et al. 2013[40] provide a detailed overview of the BiographyNet schema and the motivation behind it. We based our design on two fundamental principles: (1) maintain all information provided by the original data and (2)

---

maximize insight into the provenance of data. Insight into the provenance data is not restricted to pointing back to the original source or text. It also includes representing the process that was applied to represent the data in RDF.

The BiographyNet website provides an illustration of the overall model.[41] We base our model on the Europeana data model.[42] Persons described in the BPN are considered to be Europeana Provided Cultural Heritage Objects (edm:ProvidedCHO). Each biographical entry is seen as a specific 'view' on that person, and is linked to the person through an Aggregation object. Each biographical entry consists of three components in the original data: the file description providing information about publisher and author, the person description providing metadata about the biographee (such as name, date and place of birth and sometimes information about residence, education and occupation) and biographical materials such as images and text. The BPN stores this information in XML files. The first step in creating our structured data for the BiographyNet demonstrator was to convert these XML files to RDF. The schema used for this conversion preserved the original XML structure as much as possible.

The enrichments created by the NLP pipeline involve both person description data and textual biographical elements. The output of our analyses creates a new biographical entry linked to the edm:ProvidedCHO object representing the person. This new biographical entry contains new person metadata resulting from the NLP process. A new biographical entry is added for this, in order not to compromise the original metadata in case of conflicting information. However the general structure of the person metadata is the same as for the original entry. The details of this structure will be described in the next subsection.

An essential part of representing this data is provenance modeling. As pointed out at various stages in this paper, the historian must have all possible means available to assess the reliability of information. It is therefore essential that the historian knows where specific data came from (what source), how it was created (is it converted metadata or the result of automatic text analysis) and what potential influence the creation process had on the reliability of the data. Insight in the process of creating data can also be of importance for computer scientists and computational linguists responsible for creating the tools and data. Detailed information about how data was created can help to replicate results, spot errors in the process and improve the used tools.

---

41  http://www.biographynet.nl/wp-content/uploads/2013/07/BiographyNet_schema_with_provenance_ V1_2.png

42  Martin Doerr / Stefan Gradmann / Steffen Hennicke / Antoine Isaac / Carlo Meghini / Herbert van de Sompel, The Ruropeana Data Model (EDM), In: *World Library and Information Congress: 76th IFLA general conference and assembly*, 2010 ,10–15.

We provide this insight by modeling provenance using the W3C recommended PROV-DM[43] combined with P-PLAN, which can be used to represent the planned processes.[44] PROV supports three views in provenance modeling: The data view allows us to model which sources specific data was derived from. In this view, we link extracted information to the original text, for instance. The process view allows us to represent the processes used to create the data. In this view, we represent the tools (including their version) that were involved in converting the original XML structure to RDF or in processing the biographical text to extract information. The responsibility view provides the means to represent the agents involved in setting up or executing the processes. This ultimately leads to contact information of the developers of the pipeline. The modeling of plans using P-PLAN is set up parallel to the overview of processes. These plans model what is supposed to happen at each step, and can be used to verify whether individual steps were carried out according to plan. As such, they can serve an important role in debugging the data creation process.

Historians using the data and computational linguists or computer scientists interested in replicating data or setting up comparable systems do not require the same information. Historians must be able to go back to the original source to verify whether the information provided by the structured data correspond to that provided by the original text. They furthermore need to know whether data came directly from the metadata or involved automatic analyses and, in the latter case, they would ideally have access to a comprehensible description of how the tools used work and what biases they might introduce. Computer scientists or computational linguists need detailed information about the exact programs that were used, ideally consisting of links to repositories accompanied by the exact version that was used to create this particular dataset.

Our provenance representation fulfils these goals by modeling provenance at different levels. At an aggregated level, each person description is related to its origin: either the metadata from the original source or the text that was analyzed by the NLP pipeline. We also indicate the exact tokens that provided certain information together with their offsets. Historians can thus quickly access original information and use this to verify whether statements in structured data are correct. In addition, our representation contains a reference to the overall process accompanied with a link to documentation oriented at the end users. For computational linguists and computer scientists, we represent each step in the process using links to github repositories, references to the official version of the tool as well the exact commit that was used in the process. This allows them to recreate the exact pipeline that was used to create a specific dataset.

### Person Descriptions in BiographyNet

In addition to the name of the person, their gender and occasionally parents or partners, the metadata can provide various events and states in the biographees life. Their birth, death and marriage are represented as events (with dates and locations) and education and occupation as states. As mentioned above, we stay as close as possible to the original structure when converting metadata to RDF. Person descriptions are thus linked to events for which we can define a type (e.g. *birth, death*), a date and a location (if given). As such the original XML structure is directly compatible with the Simple Event Model (SEM).[45] SEM is a highly flexible model that can be used to represent events, their participants, location and time. Because of its flexibility and compatibility with the data representation already present in the BPN XML data, we use SEM to represent the events identified by the NLP pipeline.

Figure 3 provides a reduced example of how an extracted event is represented in the BiographyNet datamodel. The blue circles represent the biographical components. The top circle in the middle is a new Biographical Description specifically created to represent the biographical information extracted by the pipeline. It is a proxy in the person description (the blue circle far left) and part of an aggregated representation of an individual, represented by its BNP identifier. The new biographical description is derived from the biographical elements (containing the text) of a biographical entry from the BNP. This is indicated by the *prov:wasDerivedFrom* relation. We also link it back to an identifier representing the exact process that was used to extract the data (using

---

45   Willem Robert van Hage / Véronique Malaisé / Roxane Segers / Laura Hollink / Guus Schreiber, Design and use of the Simple Event Model (SEM), in: *Web Semantics: Science, Services and Agents on the World Wide Web* 9, no. 2, 2011, 128–136.

*prov:wasGeneratedBy*). This process provides access to information about the individual tools that were used.

The content that is extracted is linked to the person description using the *bgn:Includes* relation, specifically designed for the project. This relation links a person description to all events and concepts mentioned in the biography. It is used for studies involving questions of where certain concepts are mentioned. The two lilac circles represent *instances*; the one on the left is an event instance and the one on the right represents a person. The event instance is linked to interpretations of the event (marked in green): a wordnet identifier on the top, the FrameNet frame for marriage and *sem:Event* which indicates that we are dealing with some event. The person is linked to the event with a *propBank:Arg0* relation indicating that the person was the agent of the event. Both instances are linked to their *mentions*, tokens in the original text, through the *gaf:denotedBy relation*. Mentions give additional information of their location in the text (through the indices), the lemma of the chosen words and additional linguistic information. Our example indicates that the expression referring to the person is a pronoun.

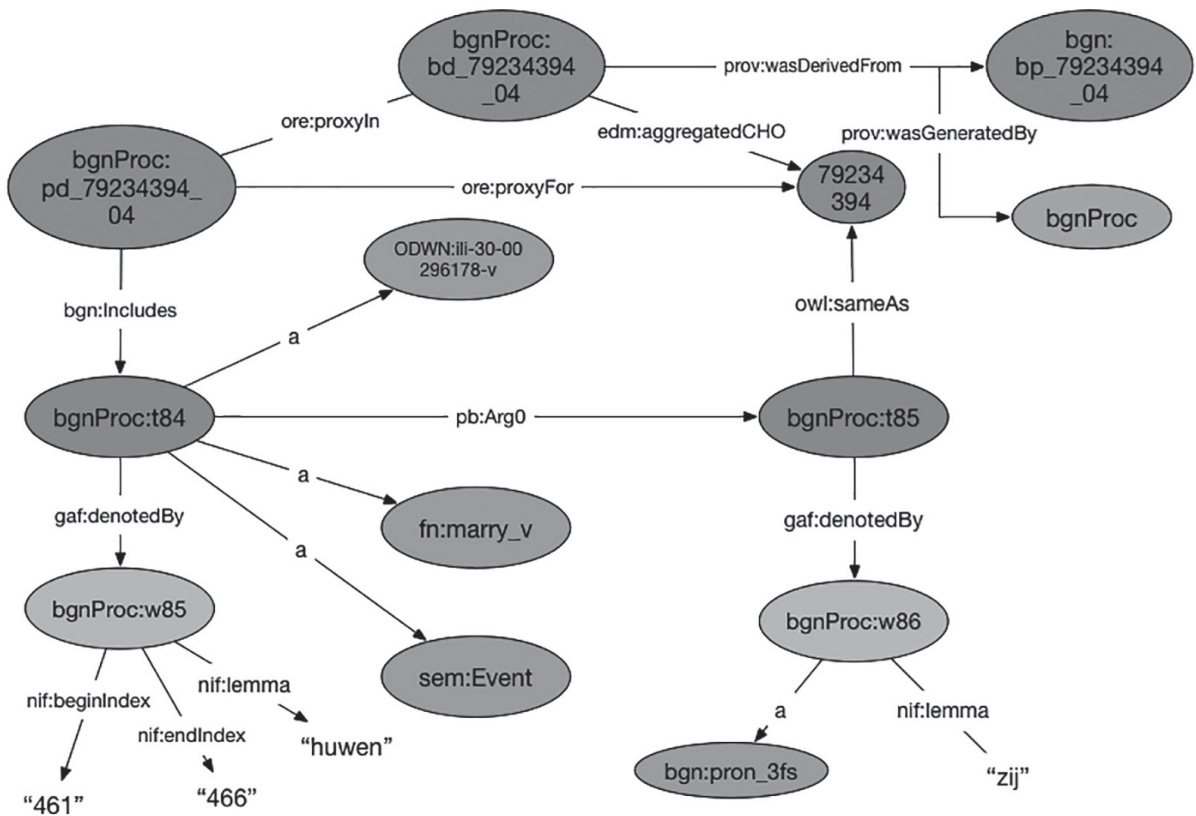

Fig. 3: example of event representation in BiographyNet

Finally, in our interpretation step, we link instances to their referent where possible. If a name corresponds to that of the biographee or there is a pronoun whose gender corresponds to the biographees gender, this is linked to the representation of the person. This representation tells us, that the biography mentioned a marriage, expressed by a token with the lemma *huwen* starting at index 461, where the person who got married is the biographee (*owl:sameAs* relation from the person instance to the edm:ProvidedCHO representation of the person), which was expressed by a feminine pronoun.

Historians can extract which lemmas, which wordnet concepts or which FrameNet frames were mentioned in various biographies. They can also perform targeted searches on specific events: if there was a marriage mention, was it the biographee marrying and who did they marry? Time references and locations are attached to events in similar manners as participants. It is thus also possible to extract timelines of a person's life or investigate where they have been. Finally, provenance information is given at two levels: first, we explicitly model where each piece of extracted information was mentioned. Historians have access to the lemmas, but we can also represent the source text and highlight the mentions in this text by using the indices provided with the mentions. At the same time, we point back to the process that was used to create a specific dataset. Here, the historian will find more information about the tools and links to tool documentation.

## 7. The demonstrator and Visualizations

The data model described in the previous section can be queried using SPARQL. SPARQL is the W3C recommended query language for RDF.[46] However, we can not expect all historians interested in using BiographyNet data to be SPARQL experts. We are therefore developing a *demonstrator* that provides intuitive access to the data as well as additional visualizations to support research on this data.

As a first step, a user study was conducted to investigate which features were considered important to historians.[47] This study based on interviews revealed that historians assign high importance to the possibility of going back to the

---

original source in order to check results. Furthermore, they appreciate suggestions from the interface about potential relations between people, would ideally have access to multiple visualizations and want to be able to store the results of their study in text form or some other human readable format, in order to use results separately from the tool. Some users also indicated interest in statistical information, insight in how the tools work and replicability of the study, though these requirements were not shared by all participants. On designing prototypes of our demonstrator and visualizations these insights were taken into account. In this section, we describe the design of the demonstrator prototype as well as an additional visualization of biographical data.

### BiographyNet Demonstrator

The BiographyNet demonstrator will provide access to the biographical data and enrichments resulting from the NLP process through various visualizations and data views. These can be broadly divided into four categories: timeline views, map views, graphs views and raw data views. These different visualizations exploit the main data elements within the biographical data: time periods, locations and relations of people, places and events. As such, each type of visualization can provide a different insight. In order to access the data, the demonstrator contains a general search bar as found in major web search engines. This text based search functionality uses information from metadata as well as full biographical descriptions. After the initial search, relevant persons are shown in the chosen visualization, one of which can be chosen using tabs. The initial result can then be refined based on a collection of metadata fields derived from the result. Hence, a combination of free search and faceted browsing is used. The results of different searches can be viewed together and combined in order to compare them. Meanwhile, the demonstrator keeps track of all these search and refinement operations allowing the historian to either go steps back in the process, branch off in a different direction, or store these steps in order to later reproduce the same results.

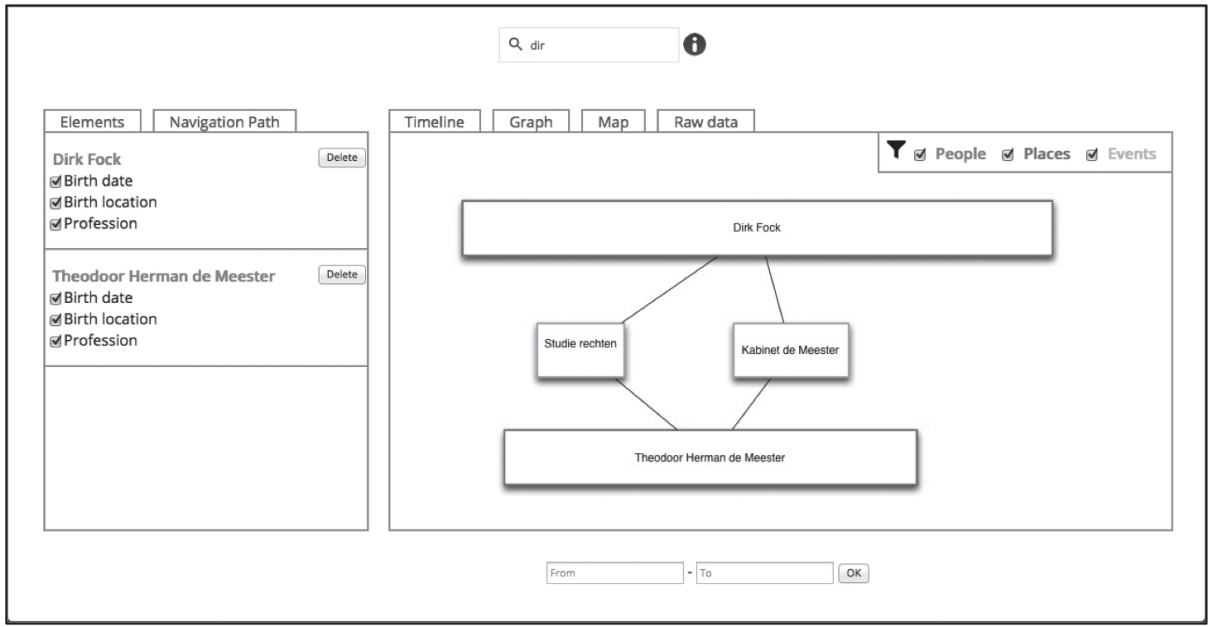

Fig. 4: Mock up of the demonstrator's main screen showing the general search bar, timeline visualization and metadata-based refinement display.

By design, the demonstrator should provide information to the historian in an insightful way, but leave the interpretation of that information—and more importantly—the judgement of it, to the historian. The provenance information stored in the BiographyNet data model will be used to facilitate this. It therefore plays an important role in establishing the credibility of the demonstrator as an academic research tool. Though there are general purpose visualization solutions for provenance information available such as Prov-O-Viz,[48] these require a deeper understanding of the provenance model, due to the lack of context. In a purpose-built visualization, this context can be taken into account; the provenance information can be displayed based on the well-understood principle of having multiple biographical descriptions per person. Furthermore, the provenance view can be combined with "related but non-prov" data such as token offsets stored by the NLP pipeline in order to highlight the exact text phrases used by the NLP, next to the actual provenance data on that particular process.

When a metadata fact is selected in a visualization, e.g. the birthdate of a person is selected by clicking on the beginning of that person's timeline, a screen is displayed that provides the selected fact, possible alternatives for it

that may be contained in alternative biographical entries, and an overview of
the aggregated provenance information for those facts. By clicking on a meta-
data fact, a fold-out view is given of the sources that agree with this fact, contra-
dict with it, or partially confirm it. Using this display, the historian can see
which sources, i.e. which biographies, were used to derive that particular fact.
Furthermore, a fragment from the actual biography can be displayed, high-
lighting the exact terms that were used. This approach does not only allow the
historian to directly check the fact in the original source, but also provides a
form of context.

Fig. 5: Mock up of the demonstrator's fact info screen showing several alternative birth
metadata facts, the sources overview for one of those facts, and the original source
texts.

The actual BiographyNet demonstrator is currently under development and will be based on these prototypes. The interface of the BiographyNet demonstrator will be created using Data 2 Documents,[49] a vocabulary for creating documents out of data, developed under auspices of the BiographyNet project. Using the Data 2 Documents vocabulary, definitions can be created that determine how the (meta)data related to a particular data resource should be rendered in HTML in specific situations, which fits well with the rendering of metadata in different views for the demonstrator interface.

### BiographyNet visualizations in Storyteller

Storyteller aims at visualizing events that are structured as stories. The visualization is developed based on three projects: NewsReader,[50] Embodied Emotions[51] and BiographyNet. Despite the difference in the data and goals of this project, Storyteller provides means to gain insight of the data in all three projects. We will limit our description to the BiographyNet visualizations in this paper.[52]

Storyteller provides three views on the data. The first view is actor-centric. This view provides so-called participation graphs. The x-axis of this representation is a timeline and the y-axis provides a list of actors. Each actor is represented by a line in its own color. The events the actor is involved in are represented as circles on these lines. If two actors are involved in the same event, the lines intersect at this event. In the BiographyNet visualization, we represent all events of the same type occurring in the same year as a single event, i.e. we represent the births, marriages, deaths, etc. occurring in the same year together. People's life-lines intersect when they are born in the same year, get married in the same year, etc. Left of these life-lines an alphabetic list of the actors is given together with a horizontal bar in the participant's color. The length of this bar indicates the number of events the actor is involved in.

---

The second view provides two event-centric visualizations of scattered events on timelines. Events are classified into groups. The first visualization represents all events of a given group on a single row. Events receive a climax score indicating their importance. The group containing the event with the highest climax score is presented in the top row, followed by the group with the second highest climax score, etc. In the second visualization, each individual event is placed on the y-axis depending on its own climax score. Events belonging to the same group are thus placed on different rows. Events belonging to the same group are represented by the same symbol and color. This view thus allows the user to see how the climax scores of a given event group change over time. If all events belong to the same story (e.g. by common actors), this visualization shows how the story unfolds into a climax and how the aftermath slowly fades out. In the BiographyNet visualization, events of the same type belong to the same group and the climax score is determined by how many people participated in this type of event in a given year. This visualization does not therefore tell the story of individual people, but rather of types of events revealing peaks in births, marriages, studies, deaths, etc. in a specific dataset.

Finally, the third view fulfils the verification requirement of the historians. This view provides sentences from the original data highlighting the place where individual events are mentioned. Historians can see in this view whether automatic analysis identified the correct event. If no sentence is provided, the user knows this data point came from the metadata. Overall, the visualization points out who experienced the same (kind of) thing at the same time, supporting analyses on what happened to people who were born or died at a specific time, who studied in the same year etc. As such, the BiographyNet Storyteller supports generation-based investigation.

## 8. Conclusion, Discussion and Future work

This paper described the BiographyNet project. BiographyNet combines expertise from the fields of history, computational linguistics and computer science to create order in a mass of unstructured, or semi-structured, biographical data and create new insight by providing links to previously separate data. Historical research drove the questions which the tool should be able to answer, computational linguistics wrote the algorithms to automatically analyze the texts and computer science converted the data into linked data and designed a user friendly and inspiring interface. All fields were equally re-presented, which we feel is a necessary condition for making such a project successful.

We started with unstructured short biographies on circa 75,000 famous people from Biography Portal of the Netherlands (BNP). Some metadata fields were already added to their biographies, but some information like 'parents', 'occupation' and 'education' were frequently missing. Often these elements were present in the text and we used the BiographyNet pipeline to extract them and enrich the metadata of the text. Our demonstrator aimed at visualizing this data in a user friendly manner, providing leads rather than answers and paying attention to provenance and different perspectives.

Provenance and perspectives were the two themes that bound the three fields together when creating the demonstrator. The historian always wants to know what the original source of a piece of information is, to verify what is stated. Sometimes data will conflict, and sometimes there is a different perspective on a person or event. Not only the data itself has to deal with provenance and perspectives, but also the NLP enrichments. For every enrichment, provenance is specifically defined including original source and a reference to how the enrichment was created. The different perspectives on people through time which can thus be extracted form an interesting research project in itself. The demonstrator is planned to be published in the Fall of 2017.

The evaluation of the NLP pipeline is currently ongoing. We have argued that such an evaluation should both include intrinsic evaluations and extrinsic evaluations. Intrinsic evaluations should give an indication of the performance of individual NLP tools on the data from the biography portal. Extrinsic evaluation investigates the suitability of the generated data to address specific historical questions. This evaluation is not merely interested in the numbers, but also in the kind of errors that occur and whether they introduce a bias influencing the outcome of the historian's question. Though we consider the requirements for evaluations one of the main important insights gained within this project, it also forms one of the most important lessons learned in how to plan a digital humanities project. Rather than running an evaluation at the end of a project, these projects should start by setting up evaluations.

BiographyNet resulted in numerous publications and national and international collaborations. A lot more, however, could still be done on this topic. There are biographical dictionaries worldwide, which could be analyzed using the same techniques and methodology. Not only could these dictionaries and other biographical resources be enriched the same way as the BNP, but it is also possible to link these different datasets with one another for analysis. Questions regarding world renowned people like Erasmus could be addressed while using biographical datasets in different languages and from different countries. Conflicting data and different perspectives will surface. A general picture on the good and bad qualities ascribed to people could provide us with insights into European culture and even into nationalism and state formation in the nineteenth and twentieth centuries. Finally, trends can be validated across national borders and language-communities.